\documentclass[runningheads,a4paper]{llncs}
\usepackage{graphicx}
\usepackage[tight]{subfigure}
\usepackage{float}
\usepackage[ruled]{algorithm2e}
\usepackage[usenames,dvipsnames]{xcolor}
\usepackage{amssymb,amsmath}
\usepackage{url}

\begin{document}
\subfiglabelskip=1pt
\subfigcapskip=0pt
\subfigcaptopadj=0pt
\subfigbottomskip=0pt
\subfigcapmargin=1pt

\author{
Jieqing Jiao\inst{1}
\and S{\'e}bastien Ourselin\inst{1,2}
}
\index{Jiao, Jieqing} 
\index{Ourselin, S{\'e}bastien} 

\institute{Translational Imaging Group, Centre for Medical Image Computing, UCL, UK \and
Dementia Research Centre, Institute of Neurology, University College London, Queen Square, WC1N 3BG, UK \\
\email{\{j.jiao, s.ourselin\}@ucl.ac.uk}
}

%
\mainmatter  
\title{Fast PET reconstruction using Multi-scale Fully Convolutional Neural Networks}
\titlerunning{Multi-scale Fully Convolutional Neural Networks (msfCNN)}

\maketitle
\begin{abstract} 
Reconstruction of PET images is an ill-posed inverse problem and often requires iterative algorithms to achieve good image quality for reliable clinical use in practice, at huge computational costs. In this paper, we consider the PET reconstruction a dense prediction problem where the large scale contextual information is essential, and propose a novel architecture of multi-scale fully convolutional neural networks (msfCNN) for fast PET image reconstruction. The proposed msfCNN gains large receptive fields with both memory and computational efficiency, by using a downscaling-upscaling structure and dilated convolutions. Instead of pooling and deconvolution, we propose to use the periodic shuffling operation from sub-pixel convolution and its inverse to scale the size of feature maps without losing resolution. Residual connections were added to improve training. We trained the proposed msfCNN model with simulated data, and applied it to clinical PET data acquired on a Siemens mMR scanner. The results  from real oncological and neurodegenerative cases show that the proposed msfCNN-based reconstruction outperforms the iterative approaches in terms of computational time while achieving comparable image quality for quantification. The proposed msfCNN model can be applied to other dense prediction tasks, and fast msfCNN-based PET reconstruction could facilitate the potential use of molecular imaging in interventional/surgical procedures, where cancer surgery can particularly benefit.

\end{abstract}

\section{Introduction}

Positron Emission Tomography (PET) provides non-invasive \textit{in vivo} imaging of radioactive tracer that binds to a particular molecular target of interest. It is by far the most sensitive modality for non-invasive molecular assaying of the human body, and plays an important role in understanding and diagnosis of a wide range of diseases. A PET scanner collects photons emitted from radioactive tracer molecules serving as the imaging agent, and the reconstruction of PET images is the estimation of  tracer distribution, which can be used for quantification of the molecular target of interest. PET reconstruction can be done by analytic methods to directly calculate the image in simple steps, such as filtered back projection (FBP), which has been commonly used in computed tomography (CT) reconstruction. However  PET reconstruction with analytic methods is very challenging due to the fact that, in PET imaging, the emission of photons is a stochastic process and the noise statistics are hard to model by the analytic methods. For this reason, PET reconstruction algorithms based on a statistical model in the maximum likelihood (ML) framework have been developed with the advantages of having high flexibility in modelling the data acquisition process \cite{Jiao2015}. These methods usually require an iterative optimisation approach to keep updating the reconstructed image to fit the measured photon data, and can be very computationally expensive as the flexibility/complexity of the model grows for improving reconstruction image quality. The long reconstruction time required by the iterative methods also impedes the opportunities of incorporating PET imaging into interventional/surgical procedures, where cancer surgery can particularly benefit, for example, from in-operation tumour removal assessment \cite{Solomon2016}.

Recently, with the development of deep learning, neural network based methods have shown promise in providing exciting solutions for medical/biological imaging problems \cite{Ronneberger2015}\cite{Oktay2016}\cite{Wuerfl2016}.  The flexibility based on the universal approximation rule allows deep neural networks to learn complex nonlinear relationships in a task while being computationally simple.

In this work, we aimed to address the speed-quality trade-off in the existing PET reconstruction methods (analytic and iterative) by proposing an alternative approach for improvements in both accuracy and computational performance.  We consider image reconstruction a dense prediction problem, for which solutions based on deep neural networks are rapidly developing, and we proposed a novel multi-scale fully convolutional neural network (msfCNN) for learning contextual features with both memory and computational efficiency.  The  model was trained using simulated data and then applied to real clinical data for fast reconstruction. The experimental results show that the msfCNN-based method can effectively reduce the reconstruction time while achieving comparable image quality for target density quantification and tumour detection.

\section{Method}

\subsection{PET reconstruction as a dense prediction problem}
Image reconstruction can be considered as a dense prediction problem, where for each voxel/pixel, a model predicts a value in the reconstructed image from the input data. In PET imaging, the raw data from the scanner are photon emission counts, which can be arranged as sinograms in the projection space. The mapping from the projection space to the image space can be achieved by back projection, which is a linear transform. In~\cite{Wuerfl2016} the back projection was implemented by a fully connected layer in the neural network, however this can introduce millions of parameters, since the number of parameters $n_{p} = n_{v}\times n_{d}$, where $n_{v}$ and $n_{d}$ are the numbers of voxels and photon detector pairs respectively. The back projection can also be calculated by inverse radon transformation with no need for parameters, where the measurement uncertainties are transferred into the image space, resulting in radial streak artefacts and high noise which are challenging to remove while preserving the original resolution. In this work, we firstly used inverse radon transformation to project backwards the photon counts data into the image space, and focused on reconstructing the PET images from there. The goal was to design a neural network with the capacity to compensate for the uncertainties in the photon emissions, which make PET reconstruction an ill-posed inverse problem, by using the large scale contextual information in the back projection image to accurately recognise and remove the artefacts and noise, hereby recovering the original image of the tracer distribution. We designed a novel multi-scale fully convolutional neural network (msfCNN) which achieves large receptive fields with both memory and computational efficiency, and does not suffer from spatial resolution loss while down-scaling and up-scaling, leading to less blurry reconstructed PET images.

\subsection{Architecture of the multi-scale fully CNN (msfCNN)}
\begin{figure}
\centerline{
\includegraphics[width=1.0\textwidth]{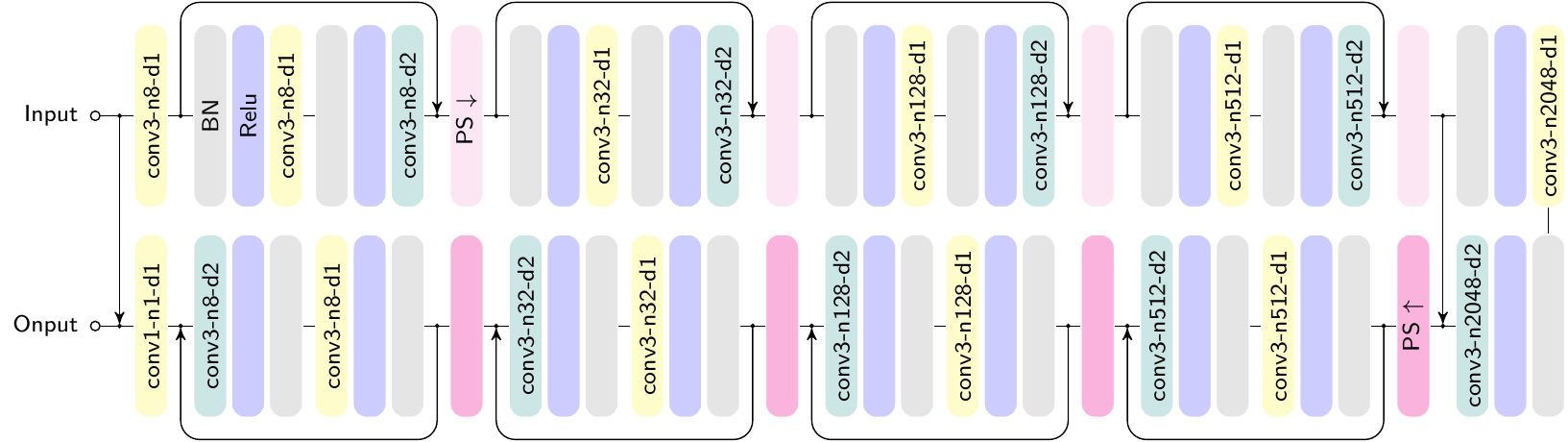}
}
\caption{Architecture of the proposed multi-scale fully convolutional neural network (msfCNN) for 2D. For PET reconstruction the input is the back projection of photon emission data (in the image space) and the output is the reconstructed PET image. The numbers in the convolutional layers correspond to the size of kernels, the number of feature maps (n) and the dilation rate (d) for dilated convolution. The model has a downscaling-upscaling structure without resolution loss by using the periodical shuffling (PS) operations. The PS $\uparrow$ operations have the same definition as in sub-pixel convolution to reshape the feature maps by a factor of $2\times2$ in upscaling, and PS $\downarrow$ is its inverse operation in downscaling, and the total feature map size ($height\times  width \times depth$) does not change throughout all the layers. Arrows denote element-wise sum for residual connections.}
\label{fig:}
\end{figure}

In this work, we constructed the network in two dimensions. The proposed multi-scale fully convolutional neural network (msfCNN) consists of convolutional layers with the kernel size of $3\times3$. Efficient receptive field increase was obtained by incoporating dilated convolutions \cite{Yu2015} which can aggregate multi-scale contextual information with the same number of parameters. All dilated convolutions have the dilation factor of $2$. To reduce the memory costs from having a constant size ($height\times width $) for all the feature maps throughout the network, previously pooling was used to downscale the spatial size of feature maps gradually as the network grows deeper with the increasing number of channels, however this can result in significant resolution loss which will be hard (or impossible) to recover by later layers. In \cite{Shi2016} sub-pixel convolution with a periodic shuffling operator was proposed to jointly learn  the feature extraction and up-sampling filter weights for super-resolution reconstruction, where the periodic shuffling operator $PS\uparrow(T)$ is defined as, for a tensor $T$ of shape $H\times W\times r^2\cdot D$,
$$
PS\uparrow(T)_{x,y,d} =T_{\left\lfloor x/r \right\rfloor, \left\lfloor y/r \right\rfloor, D\cdot [\bmod(y, r)\cdot r+\bmod(x, r)]+d}.
$$
Inspired by this idea we derived its inverse operator $PS\downarrow(T)_{x,y,d}$, for a tensor $T$ of shape $r\cdot H\times r\cdot W\times D$, as
$$
PS\downarrow(T)_{x,y,d} = T_{x\cdot r +\bmod( \left\lfloor d/D\right\rfloor,r), y\cdot r+\left\lfloor\left\lfloor d/D\right\rfloor/r\right\rfloor,\bmod(d,D)},
$$
which allows the joint learning of the feature extraction and down-sampling filter weights when combined with the convolutional layers. As $PS\downarrow$ changes the shape of the feature maps  from $r\cdot H\times r\cdot W\times D$ to $H\times W\times r^2\cdot D$, no pooling is required, and all the information will be used by the following layers with no loss of spatial resolution. In addition, by using $PS \uparrow$ and $PS\downarrow$, the total size of the feature maps remains the same throughout all the layers which provides memory efficiency for layers with a larger number of feature maps. In this work we used the shuffling rate $r = 2$. Therefore the proposed msfCNN shares a similar encoder-decoder style structure as \cite{Ronneberger2015} but eliminates the problem of resolution loss. To facilitate training, residual connections were added, following the arrangement proposed in \cite{He2016}.

\subsection{Implementation and training  details}
The training data were generated by simulations, where an image serving as the ground truth is forward projected into the sinogram space by radon transformation, and after Poisson is added, the sinogram is backward projected into the image space by the inverse radon transformation with ramp filter and linear interpolation to generate the input image. The ground truth images in this work were extracted from  brain scans in the Alzheimer's Disease Neuroimaging Initiative (ADNI) database. The training data were generated from the axial slices from T1-weighted MR, FDG PET and AV45 PET data, and $45$k images (MR$:$FDG$:$AV45 $=$ $1$$:$$1$$:$$1$) were used for training. 
The proposed model was implemented using Tensorflow \cite{Abadi2016} with Keras \cite{chollet2015keras}, and trained with the initialisation proposed in \cite{He2015} and Adam optimiser \cite{Kingma2014} with the settings $learning$ $rate = 0.001$, $\beta_1=0.9$, $\beta_2=0.999$, $\epsilon=1e-08$, $decay=0.0$, on a Tesla K40c GPU. Mean squared error was used as the training loss function. 

\section{Results}
\subsection{Reconstruction evaluation}
 We applied the model learned from simulated data to reconstruct PET images from clinical data acquired on a Siemens Biograph mMR scanner from patients. In the absence of ground truth, we compared the performance of the proposed msfCNN reconstruction with ordered subset expectation maximisation (OSEM), which is an iterative method, provided by the manufacturer with default settings which includes a post-reconstruction Gaussian filter of $3.5 mm$. For this PET-MR scanner, the PET reconstruction uses a pseudo CT image synthesised from T1-weighted and T2-weighted MR images acquired in the same imaging session to calculate a $\mu$ map~\cite{Burgos2015} for attenuation/scatter correction. Attenuation, scatter, random, dead-time corrections and normalisation were applied to the PET data using the manufacturer's software before reconstruction. The scans were conducted in dynamic mode, and the reconstruction was performed to produce a series of time frames. PET data were acquired in list mode in 3D, and for the proposed msfCNN, which was implemented in 2D, the 3D projection data were firstly converted into 2D in the axial direction by using single-slice rebinning and then projected backwards by inverse Radon transformation with ramp filter and linear interpolation.

\subsection{Amyloid imaging} 
\begin{figure}
\includegraphics[width=\textwidth]{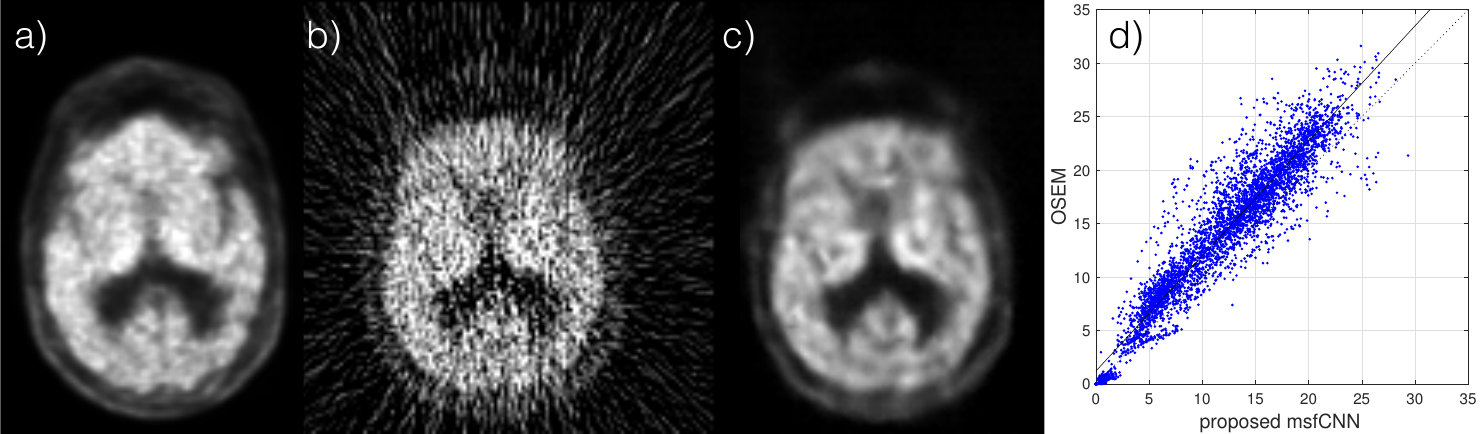}
\caption{ [\textsuperscript{18}F]florbetapir PET reconstruction for neurodegenerative imaging. a) One slice from the OSEM reconstruction using the manufacturer's software with default settings and $3.5mm$ Gaussian smoothing (unseen by the proposed model). b) Backprojection of PET sinogram data served as the input for the proposed model. c) Reconstructed slice by msfCNN. d) Plot of regional uptake values from 158 regions  $\times$ 29 time frames (correlation coefficient $r = 0.9503$), with the identity line  and  regression line.}
\label{fig:AV}
\end{figure}

Figure \ref{fig:AV} shows the reconstruction of clinical [\textsuperscript{18}F]florbetapir data of a patient with Alzheimer's disease (AD). [\textsuperscript{18}F]florbetapir is a PET radiotracer that binds to amyloid-$\beta$, which is considered to be a major target in the AD brain. Regional uptake of the tracer is of interest to assess the patient's amyloid-$\beta$ load, and the values were extracted from the reconstructed images based on the brain parcellation generated by using GIF \cite{Cardoso2015} on the subject's T1-weighted MR image. The regional uptake values are shown in Figure \ref{fig:AV} for 158 brain regions.  

\subsection{Tumour imaging} 
\begin{figure}
\centering
\subfigure[OSEM reconstruction (3D)]{
\includegraphics[width=0.85\textwidth]{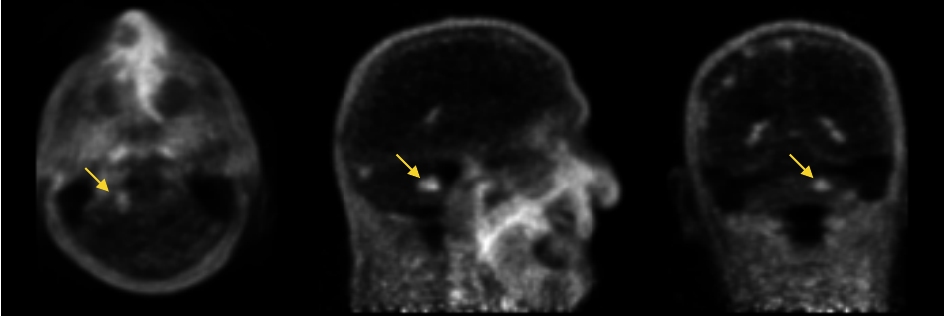}
}
\subfigure[Input for the proposed method (only the 2D axial slices)]{
\includegraphics[width=0.85\textwidth]{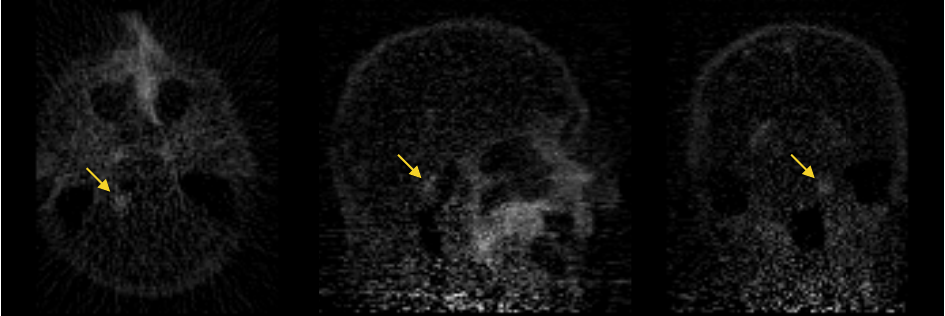}
}
\subfigure[Proposed msfCNN reconstruction (only operated on the 2D axial slices)]{
\includegraphics[width=0.85\textwidth]{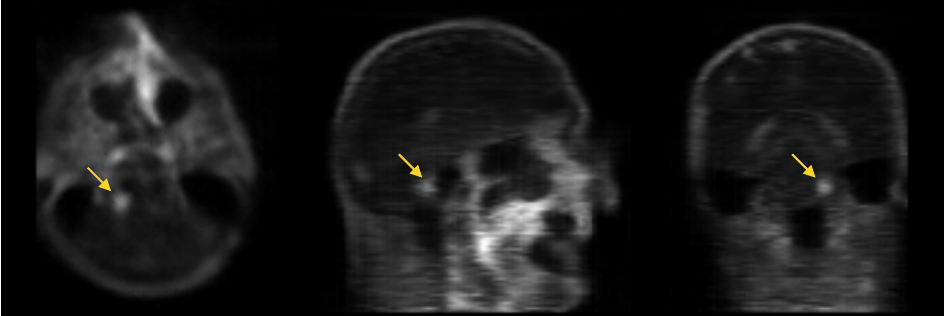}
}
\caption{Orthographic views from the reconstructions of [\textsuperscript{18}F]choline data showing a brain tumour (yellow arrows). The OSEM reconstructions were not used for training. For the proposed msfCNN method, the presence of the 'stacking' artefacts in the coronal and sagittal planes is due to the fact that msfCNN only operates in 2D on the axial slices.}
\label{fig:choline}
\end{figure}

Figure \ref{fig:choline} shows the reconstruction of a [\textsuperscript{18}F]choline scan from a patient with a brain tumour. [\textsuperscript{18}F]choline is a PET radiotracer that gets absorbed by cancer cells and participates in cell proliferation. The proposed msfCNN-based reconstruction effectively recovered the tumour for detection.

\subsection{Computational time} 
The proposed msfCNN method took $\sim$15 milliseconds for a slice, $\sim$1.9 seconds per frame and finished the reconstruction in 54 seconds for the whole [\textsuperscript{18}F]florbetapir scan (29 time frames), and 42 seconds for the whole [\textsuperscript{18}F]choline scan (22 time frames), using a Tesla K40c GPU. The OSEM method took $\sim$1 minute per frame (perhaps with overheads of preprocessing), using a Tesla K20c GPU. 

\section{Discussion and Conclusion}
In this work, we propose a novel architecture of multi-scale fully convolutional neural networks (msfCNN) for fast PET image reconstruction. The proposed downscaling-upscaling structure using the periodic shuffling operation from sub-pixel convolution and its inverse has shown to be effective in preserving spatial resolution. The model applied to clinical data was only trained with simulated data  without scanner modelling, which suggests the performance of the proposed model can be further improved by using more realistic training data. The input data for reconstruction was the back projection of sinograms. This avoids the introduction of a large number of parameters to map the back projection to a fully connected layer, and reduces the model's complexity. An end-to-end model that implements the back projection using convolutional layers will be explored in the future. In addition, the implementation of the proposed msfCNN model will be extended to three dimensions.
\newline

\bibliography{pp380}
\bibliographystyle{splncs} 
\end{document}